\PassOptionsToPackage{final}{graphicx}

\documentclass[applsci,article,submit,pdftex,moreauthors]{Definitions/mdpi}

\usepackage{booktabs}
\usepackage{multirow}
\usepackage{xurl}
\usepackage{float} 
\usepackage{graphicx}
\usepackage{amsmath} 
\usepackage{amssymb} 

\graphicspath{{figs/}}

\newcommand{\method}{UHD-GCN-BIQA}

\newcommand{\MissingGraphicBox}[2][]{%
  \begingroup
  \setlength{\fboxsep}{6pt}%
  \fbox{%
    \parbox[c][0.28\textheight][c]{0.9\linewidth}{%
      \centering
      \textbf{Missing figure file}\\[4pt]
      \texttt{#2}\\[6pt]
      Please upload it to the project (path correct?)%
    }%
  }%
  \endgroup
}

\newcommand{\SafeIncludeGraphics}[2][]{%
  \IfFileExists{#2}{%
    \includegraphics[#1]{#2}%
  }{%
    \MissingGraphicBox[#1]{#2}%
  }%
}

\newcommand{\SafeIncludeGraphicsTwo}[3][]{%
  \IfFileExists{#2}{%
    \includegraphics[#1]{#2}%
  }{%
    \IfFileExists{#3}{%
      \includegraphics[#1]{#3}%
    }{%
      \MissingGraphicBox[#1]{#2 \space or \space #3}%
    }%
  }%
}

\firstpage{1}
\pubvolume{1}
\issuenum{1}
\articlenumber{0}
\pubyear{2026}
\copyrightyear{2026}
\datereceived{}
\daterevised{}
\dateaccepted{}
\datepublished{}
\usepackage{etoolbox}
\preto{\abstractkeywords}{\nolinenumbers}

\Title{Ultra-High-Definition Image Quality Assessment via Graph Representation Learning}

\Author{Shaode Yu$^{1}$, Enqi Chen$^{1}$, Ming Huang$^{1}$, 
Xuemin Ren$^{1}$, Songnan Zhao$^{2}$, Zhicheng Zhang$^{3,*}$ and Qiurui Sun$^{4,*}$}
\AuthorNames{Shaode Yu; Enqi Chen; Ming Huang; Xuemin Ren; Songnan Zhao; Zhicheng Zhang; Qiurui Sun}

\address{%
$^{1}$ \quad School of Information and Communication Engineering, Communication University of China, Beijing 100024, China\\
$^{2}$ \quad College of Engineering, Northeastern University, 
Silicon Valley, San Jose, CA 95113, USA\\
$^{3}$ \quad JancsiLab, JancsiTech, Hongkong 999077, China\\
$^{4}$ \quad Center of Information \& Network Technology, Beijing Normal University, Beijing 100875, China}

\corres{Correspondence: Z.Z., and Q.S.; 
Email: zhangzhicheng13@mails.ucas.edu.cn, qiuruisun@bnu.edu.cn}

\abstract{Blind image quality assessment (BIQA) for ultra-high-definition (UHD) images remains challenging because native-resolution inference is computationally expensive, whereas aggressive resizing or isolated cropping may suppress scale-sensitive distortions and weaken the relationship between local artifacts and global scene context. This paper aims to improve UHD-BIQA by explicitly modeling the structural dependencies among sampled image regions rather than treating them as independent views, and a graph representation learning framework \method{} is proposed. The framework samples aspect-ratio-aligned patches from each UHD image, encodes them as graph nodes, and constructs a hybrid $k$-nearest-neighbor graph using spatial proximity and feature similarity. Residual graph convolution is used to propagate contextual information across regions, and gated attention pooling aggregates patch-level evidence into an image-level quality prediction. An exponential moving average normalized multi-objective loss function is adopted to stabilize the joint optimization of regression, correlation, and ranking objectives. Experiments on the UHD-IQA benchmark show that \method{} achieves PLCC $=0.7784$, SRCC $=0.8019$, and RMSE $=0.0519$, obtaining competitive correlation performance and the lowest RMSE among the compared methods. These results indicate that graph-based region relation modeling is effective for UHD image quality assessment, particularly for improving absolute quality score estimation under high-resolution visual content.}


\begin{document}
%
\section{Introduction}
\label{sec:intro}
BIQA aims to predict perceptual image quality without access to a pristine reference which is important in image acquisition, compression, transmission, enhancement, restoration, and quality monitoring \cite{zhu2023survey, yang2023progress, yu2023review}. In contrast to full-reference IQA methods \cite{wang2004ssim, zhang2011fsim, chen2016edge}, BIQA is more practical for real deployment, because reference images are often unavailable, and this reference-free setting makes BIQA a key technique for perceptual optimization in modern visual communication \cite{li2017evaluation, min2024perceptual}.

The development of BIQA approaches has progressed through several stages. Early methods mainly relied on natural scene statistics and descriptors, under the assumption that distortions perturb regular statistical patterns of natural images \cite{mittal2012brisque, mittal2013niqe, saad2012bliinds2, liu2014sseq, yu2023hybrid}. 
Later, deep convolutional neural network (CNN)-based, patch-based, and region-based methods substantially improved prediction accuracy by learning quality-aware representations directly from training samples \cite{bosse2018deepiqa, zhang2020dbcnn, yu2017shallow, yu2016cnn, ying2020paq2piq, yu2025exploring, chen2025taylor}. 
After that, Transformer-based BIQA has strengthened long-range context modeling and flexible input handling, while large-scale pretraining has improved generalization and representation quality \cite{you2021transformer, ke2021musiq, lee2023dual, kwon2024clip}. 
Beyond backbone changes, recent BIQA has begun to absorb ideas from foundation models, multi-modal learning, and reasoning-oriented perceptual assessment. 
Descriptive IQA, reinforcement-learning-assisted ranking, and vision--language models increasingly treat image quality as a richer semantic and perceptual reasoning problem rather than as pure score regression \cite{you2024depictqawild, li2025qinsight, wu2025visualqualityr1}. 
Remarkable progress has been witnessed in the BIQA field, and a large number of approaches have been designed and applied in real-world applications.

However, the UHD-BIQA task remains challenging. Native-resolution 4K images are typically too large for straightforward full-frame inference under practical graphics processing unit memory and latency budgets. Simple resizing is also problematic because it can suppress high-frequency distortions and alter the scale at which quality is perceived. 
Meanwhile, UHD degradations are often spatially non-uniform: localized artifacts may coexist with globally coherent structures, so perceptual judgment depends on both local evidence and scene-level context. 
High-resolution BIQA studies make this point explicit by showing that quality prediction is entangled with native resolution and display scale rather than preserved by a naive resize operation \cite{huang2024hriq, hosu2025iisa}. 
The Ultra-High-Definition Image Quality Assessment (UHD-IQA) benchmark and the Advances in Image Manipulation (AIM) 2024 UHD blind photo quality assessment challenge further establish that directly adapting conventional BIQA models to 4K imagery is difficult under both perceptual and efficiency constraints \cite{hosu2024uhdiqa, hosu2025aim_uhdiqa_lncs}.

Existing UHD-oriented solutions usually rely on resizing, cropping, multi-view fusion, or multi-branch designs. 
Representative challenge methods, such as UIQA, combine global aesthetics, local distortion fragments, and saliency-guided cues for UHD quality prediction \cite{sun2024assessing}, whereas a pseudo-reference dual-branch route based on super-resolution reconstruction has also been explored \cite{gu2025super}. 
In parallel, graph-based and relation-aware BIQA methods have shown the value of modeling distortion relations, non-local dependencies, adaptive graph attention, or inter-patch interactions \cite{sun2022graphiqa, jia2022nlnet, wang2024adaptive, liu2025iqg}. 
Nevertheless, conventional BIQA methods are not designed for UHD-specific resolution and distortion characteristics, and existing graph-based BIQA methods do not explicitly address the full UHD pipeline.
%

To address this gap, the UHD-BIQA task is formulated as a graph-level regression problem, and a graph representation learning (GRL) framework, \method{}, is proposed to bridge patch-level local evidence and image-level perceptual quality prediction. 
It samples an aspect-ratio-aligned grid of patches, extracts deep features for each patch, and designs a hybrid $k$-nearest-neighbor (KNN) graph to quantify the relations among sampled image patches. 
Further, residual graph convolutional network (GCN) layers are employed to propagate contextual cues, and gated attention pooling is used to aggregate patch evidence into an image-level quality score. 
Notably, for optimizing a novel multi-objective loss function, we use an exponential moving average (EMA) so that the regression, correlation, and ranking terms can be combined stably despite their different numeric scales. 
The main contributions of this work could be summarized as follows, 
\begin{enumerate}
\item UHD-BIQA is formulated as a graph-level regression problem in which an aspect-ratio-aligned set of patches is treated as a relational structure rather than as isolated crops, enabling modeling of inter-patch dependencies under a KNN graph. 
\item A graph reasoning pipeline is introduced that combines residual GCN propagation and gated attention readout for image quality prediction, and 
an EMA-based multi-objective optimization is designed to stabilize the joint optimization of regression, correlation, and ranking terms in the hybrid loss function. 
\item Extensive comparisons with conventional BIQA methods, UHD-oriented challenge methods, and graph-based methods are conducted on the UHD-IQA benchmark, and the effectiveness of the proposed framework is verified by its low prediction error. 
\end{enumerate}


%
\section{Related Work}
\label{sec:related}
%

\subsection{The definition of the UHD-BIQA problem}
Given an input image, $\mathbf{I}\in\mathbb{R}^{H\times W\times C}$ ($C=3$), and its corresponding score, $y\in\mathbb{R}$, obtained from human subjective studies, BIQA aims to learn a regression function $f_{\theta}$ such that $\hat{y}=f_{\theta}(\mathbf{I})$ approximates $y$.
In the UHD settings, directly feeding the full-resolution image into a deep model is often infeasible due to memory and compute constraints.
A common workaround is to use a set of patches $\{\mathbf{x}_i\}_{i=1}^N$ sampled from $\mathbf{I}$ and predict the image-level quality by aggregating patch-level evidence, \begin{equation}
\hat{y}=g_{\theta}\!\left(\{\phi_{\theta}(\mathbf{x}_i)\}_{i=1}^N\right),
\end{equation}
where $\phi_{\theta}$ is a patch encoder, and $g_{\theta}$ is a pooling/regression function to approximate the scores of input images. 

A key challenge in UHD-BIQA lies in the fact that distortions in UHD images are often {spatially non-uniform}. In other words, localized artifacts may occupy only a small portion of the image, yet disproportionately dominate the overall perceptual quality. Therefore, an effective UHD-BIQA method should, on the one hand, provide sufficiently dense spatial coverage to capture such localized degradations, and on the other hand, preserve global structural context to interpret these local artifacts within a coherent scene-level understanding.

\subsection{Graph neural network preliminaries}
GRL is to learn embeddings that jointly encode node attributes and graph topology. 
GCNs represent a major breakthrough by extending convolutional operations to graph-structured data via neighborhood message passing \cite{Khoshraftar2024gpl, Kipf2017gcn, zhang2019gcn}. 
In BIQA, graph formulations have been explored from multiple perspectives: distortion graph modeling \cite{sun2022graphiqa}, non-local dependency modeling \cite{jia2022nlnet}, adaptive graph attention \cite{wang2024adaptive}, and inter-patch message passing \cite{liu2025iqg}.
Recent efforts further enrich graph-based BIQA with more flexible relation modeling and feature fusion mechanisms \cite{jia2022nlnet, wang2024adaptive}.
Beyond plain GCNs, attention-based propagation \cite{velickovic2018gat} and expressive graph isomorphism networks \cite{xu2019gin} have also been used in computer vision tasks, suggesting that richer message passing may benefit IQA when sufficient data are available.

GRL models relational data as $\mathcal{G}=(\mathcal{V},\mathcal{E})$, where $\mathcal{V}$ is the node set and $\mathcal{E}$ is the edge set \cite{Khoshraftar2024gpl}.
In our setting, each patch corresponds to a node with a feature vector, and edges encode neighborhood relations.
A typical message passing layer updates node embeddings by aggregating information from neighbors.
The GCN \cite{Kipf2017gcn} computes
\begin{equation}
	\mathbf{H}^{(l+1)}=\sigma\!\left(\tilde{\mathbf{D}}^{-\frac12}\tilde{\mathbf{A}}\tilde{\mathbf{D}}^{-\frac12}\mathbf{H}^{(l)}\mathbf{W}^{(l)}\right),
\end{equation}
where $\tilde{\mathbf{A}}=\mathbf{A}+\mathbf{I}$ adds self-loops, $\tilde{\mathbf{D}}$ is the degree matrix of $\tilde{\mathbf{A}}$, $\mathbf{W}^{(l)}$ is learnable, and $\sigma(\cdot)$ is a nonlinearity.
Beyond GCNs, many inductive variants exist, such as GraphSAGE \cite{hamilton2017graphsage} and graph attention networks \cite{velickovic2018gat}, which differ in aggregation functions and attention mechanisms.
For graph-level prediction, node representations are aggregated by a readout function (sum/mean/max pooling or attention pooling), which is essential for BIQA because the final output is an image-level quality score.

\subsection{GRL-based BIQA methods}
Several GRL-based BIQA methods are designed from the perspectives of distortion graph representation, semantic dependency reasoning, and region-level interaction modeling. 
Sun $et$ $al$ focus on distortion-centered modeling which learns distortion graph representation over distortion types and distortion levels, and this kind of explicit relation modeling improves BIQA performance beyond isolated feature vectors \cite{sun2022graphiqa}. 
Jia $et$ $al$ model non-local dependencies by constructing graph nodes from superpixels and learning long-range regional interactions via a graph neural network, which provides a complementary graph construction strategy to regular patch-based modeling \cite{jia2022nlnet}.
Liu $et$ $al$ construct patch graphs and enable inter-patch message passing via GCN, and explicit information exchange across image regions is found beneficial the BIQA task \cite{liu2025iqg}. 
Wang $et$ $al$ further strengthen the region-level interaction modeling via adaptive graph attention that enhance both local information and contextual interactions by refining post-Transformer features \cite{wang2024adaptive}. 
Xu $et$ $al$ construct a spatial view-port graph and performing GCN reasoning across selected view-ports for omnidirectional image quality prediction \cite{xu2021vgcn}. 
Xie $et$ $al$ introduce an image quality knowledge graph which is aligned with explanatory text descriptions by exploring multi-modal input, and graph structures can also support more interpretable and knowledge-driven quality prediction \cite{xie2025kgmbqa}. 
These studies broaden the scope of GRL-based BIQA beyond patch regression and demonstrate the flexibility of relational modeling for perceptual quality prediction.

\subsection{UHD-oriented BIQA methods}
UHD-oriented BIQA methods have recently attracted increasing attention due to the rapid growth of high-resolution visual content. 
Sun $et$ $al$ assess UHD image quality from aesthetics, distortions, and saliency by evaluating resized images, grid-sampled local regions, and saliency-guided patches \cite{sun2024assessing}. 
Gu $et$ $al$ consider pseudo-reference modeling, where original patches and super-resolution-reconstructed pseudo references are paired to learn comparative quality cues \cite{gu2025super}. 
In the AIM 2024 challenge, grid mini-patch sampling and pyramid perception are explored to strengthen local evidence modeling under a stronger Transformer backbone \cite{hosu2025aim_uhdiqa_lncs}. 
These methods show that UHD-BIQA requires more careful sampling and feature fusion than conventional BIQA. However, most existing UHD-oriented methods still rely primarily on multi-view sampling, cropping, or multi-branch fusion, while explicit patch-to-patch graph reasoning remains less explored.


\section{The proposed \method{} framework}
\label{sec:method}
\begin{figure*}[ht]
\centering
  \SafeIncludeGraphicsTwo[width=0.85\linewidth]{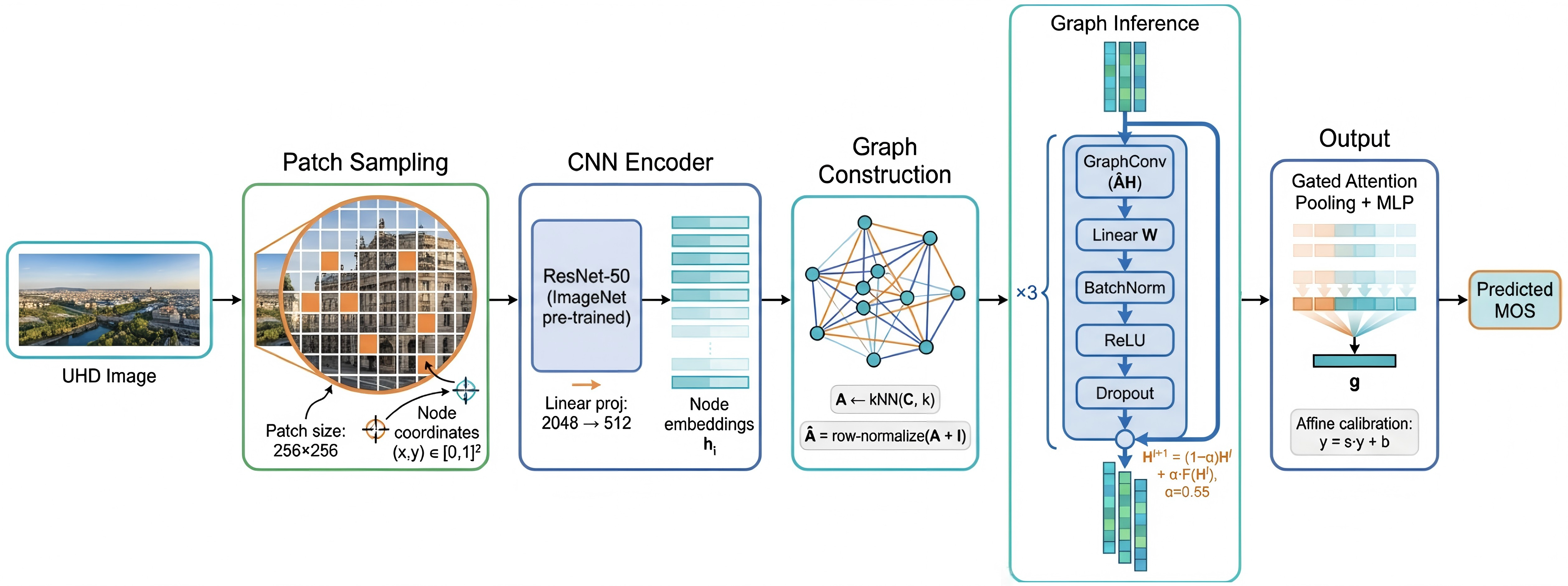}{overview.png}
  \caption{The proposed \method{} framework. A UHD image is partitioned into an aspect-ratio–aligned patch sampling from which local features are extracted. Then, a hybrid KNN graph is built to connect adjacent regions, and residual GCN layers are used to propagate contextual information. Finally, gated attention pooling aggregates the patch-level representations to produce an image-level quality prediction.}
  \label{fig:overview}
\end{figure*}

Figure~\ref{fig:overview} illustrates the proposed \method{} framework. In the framework, we sample an aspect-ratio-aligned grid of image patches, extract regional features with a CNN backbone, and build a hybrid KNN graph that combines spatial proximity and feature similarity. 
After that, residual GCN layers propagate contextual cues across the graph and a gated attention readout produces the final quality prediction. 
Finally, an EMA-normalized multi-objective function is trained to keep the optimization stable when combining regression and correlation/ranking criteria in the hybrid loss function.

\subsection{Aspect-ratio-aligned patch sampling}
Given an image $\mathbf{I}$ with width $W$ and height $H$, $N=G_w G_h$ patches of size $P\times P$ are sampled on a proper rectangular grid $(G_w,G_h)$. 
Let $(x_i, y_i)$ denote the center of the $i$-th patch in pixel coordinates. 
It is normalized to 
\begin{equation}
\mathbf{p}_i = \left[\frac{x_i}{W}, \frac{y_i}{H}\right] \in [0,1]^2 . 
\label{dftx3}
\end{equation} 
Each patch is encoded by a pre-trained backbone (ResNet-50 \cite{he2016deep}) into 
a feature vector $\mathbf{h}_i \in \mathbb{R}^d$.
Rectangular grids (e.g., $18\times 12$) align better with common UHD aspect ratios (e.g., 16:9) and yield more uniform spatial coverage under a fixed patch budget.

The key idea is to \emph{match the sampling lattice to the native aspect ratio} rather than squeezing the entire image into a square tensor.
Given a patch budget $N$, we choose a rectangular grid $(G_w,G_h)$ such that $G_w/G_h \approx W/H$, so that the spatial support of sampled patches is approximately isotropic in the \emph{image coordinate system}.
A practical heuristic is:
\begin{equation}
G_w = \mathrm{round}\!\left(\sqrt{N\cdot \frac{W}{H}}\right),\qquad
G_h = \mathrm{round}\!\left(\frac{N}{G_w}\right),
\end{equation}
followed by minor adjustments to satisfy $G_w G_h = N$.
Compared with uniform random crops, the grid ensures \emph{deterministic and uniform} coverage of the full field of view, reducing sampling variance in training and enabling fair comparison across images.
Compared with global resizing, it preserves local frequency characteristics inside each patch (no aggressive global low-pass filtering), while still providing a tractable number of tokens/nodes for subsequent graph reasoning.

We adopt a fixed patch size $P$ for two reasons.
First, it matches the input constraints of standard CNN backbones and allows us to reuse well-tested pre-trained networks (e.g., ResNet-50 \cite{he2016deep}) without architectural modification.
Second, it provides an explicit control knob for the global--local trade-off: a larger $P$ increases the receptive field per patch (more global context) but may dilute small localized artifacts, while a smaller $P$ captures fine-grained distortions at the cost of increased node count $N$ and graph complexity.
In our implementation, each patch is resized to the backbone input resolution (if needed), and we extract the global average pooled feature of dimension 2048 from ResNet-50, followed by a linear projection to $d=512$ to form node embeddings.

\subsection{Hybrid KNN graph construction}
\begin{figure}[ht]
    \centering
    \SafeIncludeGraphics[width=0.9\linewidth]{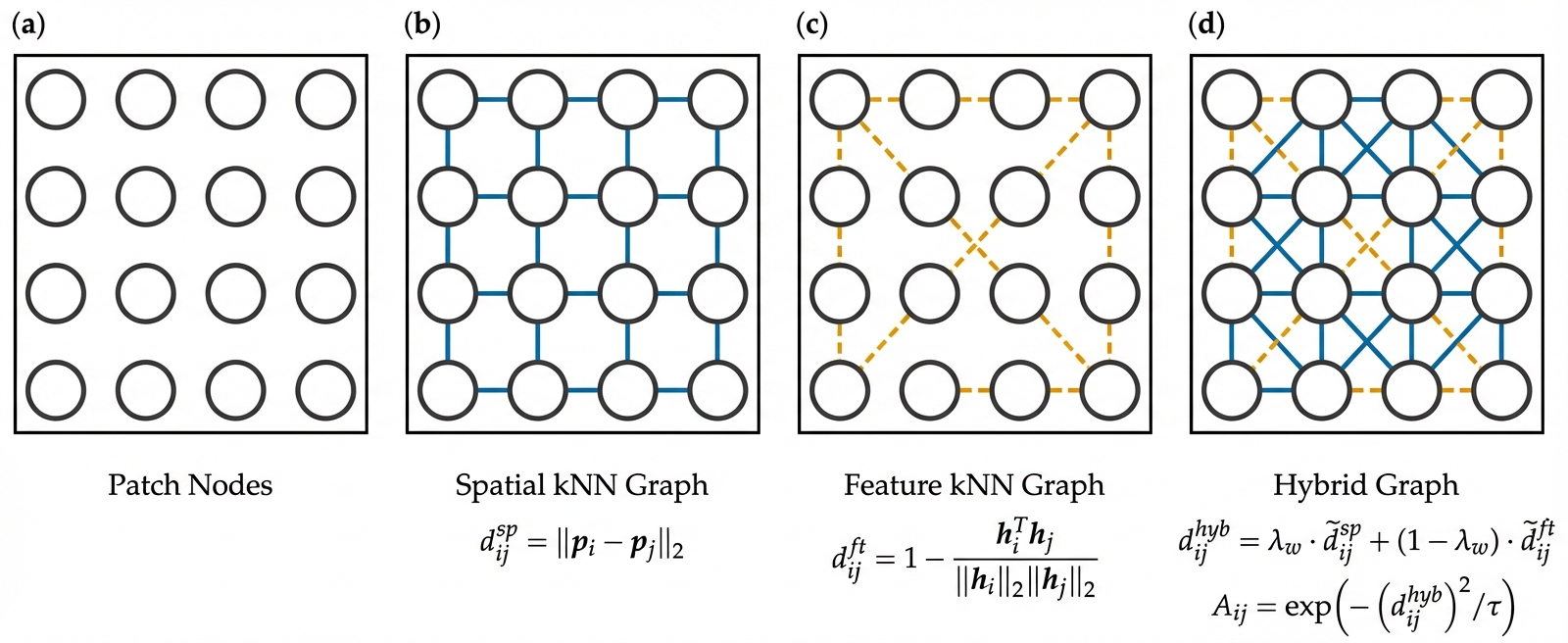}
    \caption{Hybrid KNN graph construction. 
    Each patch is treated as a node. Edges are formed based on (1) spatial proximity 
    and (2) deep-feature similarity. The two adjacency matrices are linearly combined 
    to obtain the final hybrid graph used for message passing.}
    \label{fig:hybrid_graph}
\end{figure}

Jointly considering spatial proximity and feature similarity, a hybrid KNN graph is built as shown in Figure \ref{fig:hybrid_graph}. 
Euclidean distance is defined as $d_{ij}^{sp} = ||\mathbf{p}_i - \mathbf{p}_j||_2$, 
and feature similarity is measured using 
\begin{equation}
 d_{ij}^{ft} = 1- \frac{ \mathbf{h}_i^T \mathbf{h}_j }{ ||\mathbf{h}_i||_2 ||\mathbf{h}_j||_2} . 
 \label{dftx5}
\end{equation} 

The similarities $ \{ d_{ij}^{sp} \}$ and $ \{ d_{ij}^{ft} \} $ are z-score normalized 
to generate $ \{ \tilde{d}_{ij}^{sp} \}$ and $ \{ \tilde{d}_{ij}^{ft} \} $. 
The hybrid distance metric is defined as:
\begin{equation}
d_{ij}^{hyb} = \lambda_w \tilde{d}_{ij}^{sp} + (1 - \lambda_w) \tilde{d}_{ij}^{ft},  
\label{hwimp}
\end{equation} 
where $\lambda_w \rightarrow 1$ assigns greater weight to spatial proximity. 

Edges $\mathcal{E}=\{(i,j)\mid j\in\operatorname{kNN}(i)\}$ are formed based on $d_{ij}^{hyb}$. 
The adjacency matrix weight $A_{ij}$ is computed as:
\begin{equation}
A_{ij} = \exp\left(-\frac{(d_{ij}^{hyb})^2}{\tau}\right),  
\end{equation} 
where $\tau$ is a temperature (scale) parameter controlling the bandwidth of the Gaussian affinity, which determines how rapidly the edge weights decay with respect to the hybrid distance.
This formulation follows the widely used heat-kernel-based affinity in graph learning and manifold modeling~\cite{belkin2003laplacian,coifman2006diffusion}.
In practice, $\tau$ is empirically set to 0.35, which provides a reasonable trade-off between preserving local neighborhood structures and maintaining global connectivity.

A purely spatial graph preserves geometric continuity but may fail to relate semantically similar regions that are far apart (e.g., repeated textures or symmetric structures), while a purely feature-based graph may connect such regions but ignore the strong locality prior of visual degradations.
The hybrid distance metric therefore interpolates between these two cues: $\lambda_w$ controls how strongly the graph respects geometric adjacency, and the feature term provides \emph{content-aware shortcuts} for contextual reasoning.
We z-score normalize $\{d_{ij}^{sp}\}$ and $\{d_{ij}^{ft}\}$ before mixing to prevent trivial domination by one component due to scale mismatch.
Finally, the radial basis function weighting transforms distances into soft affinities, which stabilizes message passing under noisy neighbor assignments.

In addition, we use self-loops and degree normalization to stabilize training.
In practice, we found that row-normalization (random-walk normalization) is numerically stable for large $N$ and easy to implement, while the symmetric normalization in the original GCN formulation \cite{Kipf2017gcn} can be used interchangeably.
Unless otherwise specified, we use the same normalization consistently across ablations to isolate the effect of graph construction.

\subsection{Residual GCN with controllable mixing}
Given node features $\mathbf{H}^{(l)}\in\mathbb{R}^{N\times d}$ at layer $l$, 
the graph convolution is applied:
\begin{equation}
\mathbf{U}^{(l)} = \sigma\!\left(\hat{\mathbf{A}} \mathbf{H}^{(l)} \mathbf{W}^{(l)}\right).
\end{equation}
To stabilize training and mitigate over-smoothing, residual mixing is adopted:
\begin{equation}
\mathbf{H}^{(l+1)} = \alpha \,\mathbf{U}^{(l)} + (1-\alpha)\,\mathbf{R}^{(l)}(\mathbf{H}^{(l)}),
\end{equation}
where $\alpha\in(0,1)$ controls aggregation strength.

As $L$ increases, repeated neighborhood averaging may cause node representations to become indistinguishable (over-smoothing), which is undesirable for BIQA because localized distortions must remain separable.
The mixing coefficient $\alpha$ provides an explicit control on how aggressively information is propagated: smaller $\alpha$ preserves the original patch evidence, while larger $\alpha$ strengthens contextual integration.
We set $\alpha=0.55$ in the default configuration to balance these effects.

In this study, each GCN block is implemented as a pre-activation residual unit consisting of graph convolution, linear projection, batch normalization, ReLU, and dropout.
Batch normalization stabilizes optimization under heterogeneous distortion distributions, and dropout acts as regularization against overfitting to particular patch patterns.

\subsection{Gated attention readout}
\label{sec:readout}
Node features are aggregated with a gated attention mechanism to form a graph-level representation $\mathbf{z}$.
This design is closely related to attention-based multiple-instance learning pooling \cite{ilse2018attention}, which is a natural fit for BIQA because an image can be viewed as a ``bag'' of patches where only a subset may be strongly degraded.
Concretely, we compute a scalar importance score for each node:
\begin{equation}
w_i=g(\mathbf{h}_i),
\end{equation}
where $g(\cdot)$ is a lightweight multilayer perceptron (MLP).
The attention weights are obtained by a softmax normalization:
\begin{equation}
a_i=\frac{\exp(w_i)}{\sum_{j}\exp(w_j)},
\end{equation}
and the pooled representation is
\begin{equation}
\mathbf{z}=\sum_i a_i\mathbf{h}_i.
\end{equation}
Finally, an MLP regressor $f(\cdot)$ predicts the raw quality score $\tilde{y}=f(\mathbf{z})$ and we apply an affine calibration
\begin{equation}
\hat{y}=s\tilde{y}+b,
\end{equation}
to better match the MOS scale.
Attention pooling is particularly important for UHD distortions because degradations are often spatially sparse: the model must emphasize informative regions (e.g., blocking around edges, ringing near high-contrast structures) while still integrating global context.

\subsection{EMA-normalized optimization of a hybrid loss function}
\label{sec:loss_method}
BIQA aims to produce predictions that are accurate in \emph{absolute score} (low error), consistent in \emph{linear correlation}, and reliable in \emph{ranking}.
To this end, we propose a composite objective function, 
\begin{equation}
\mathcal{L}(\theta)=\sum_{k\in\mathcal{K}} \lambda_k\,\mathcal{L}_k(\theta), 
\label{loss}
\end{equation}
where $\theta$ denotes model parameters, $\mathcal{K}=\{\mathrm{mse},\mathrm{corr},\mathrm{rank},\mathrm{var}\}$, and $\lambda_k$ are user-specified coefficients.
In practice, $\mathcal{L}_{\mathrm{mse}}$ encourages accurate regression, $\mathcal{L}_{\mathrm{corr}}$ penalizes correlation inconsistency (e.g., $1-\mathrm{PLCC}$), $\mathcal{L}_{\mathrm{rank}}$ enforces pairwise ordering constraints, and $\mathcal{L}_{\mathrm{var}}$ regularizes prediction dispersion to avoid degenerate outputs.

A known issue in multi-objective training is that different objectives may have vastly different numeric scales and gradient magnitudes.
The parameter update is driven by
\begin{equation}
\nabla_{\theta}\mathcal{L}=\sum_k \lambda_k\,\nabla_{\theta}\mathcal{L}_k,
\end{equation}
so if some $\mathcal{L}_k$ is systematically larger (or has larger gradients), it can dominate optimization even when its semantic importance is not greater.
This is particularly relevant for BIQA because ranking- or correlation-based losses may have bounded ranges, while MSE-like losses may vary widely across datasets and training stages.

We mitigate this issue by normalizing each term using a running estimate of its typical magnitude.
Let $\mu_k \approx \mathbb{E}[\mathcal{L}_k]$ denote the characteristic scale of the $k$-th loss.
We estimate $\mu_k$ online with an EMA, 
\begin{equation}
\hat{\mu}_k^{(t)}=\beta\,\hat{\mu}_k^{(t-1)}+(1-\beta)\,\mathcal{L}_k^{(t)},\qquad \beta\in(0,1),
\end{equation}
where $t$ is the iteration index. Optionally, a bias-corrected estimate can be used as $\tilde{\mu}_k^{(t)}=\hat{\mu}_k^{(t)}/(1-\beta^t)$, 
while in our experiments the uncorrected EMA is sufficient when $\beta$ is close to 1 and training runs for enough iterations.
We then define the \textbf{EMA-normalized} objective:
\begin{equation}
\mathcal{L}_{\mathrm{total}}(\theta)=\sum_{k\in\mathcal{K}} \lambda_k\,
\frac{\mathcal{L}_k(\theta)}{\mathrm{stopgrad}(\hat{\mu}_k)+\epsilon},
\label{eq:ema_norm_loss}
\end{equation}
where $\epsilon$ is a small constant for numerical stability, and $\mathrm{stopgrad}(\cdot)$ indicates that the EMA statistics are treated as constants with respect to $\theta$ (i.e., detached from back-propagation).

With in Eq. \ref{eq:ema_norm_loss}, the gradient becomes
\begin{equation}
\nabla_{\theta}\mathcal{L}_{\mathrm{total}}=
\sum_{k\in\mathcal{K}}\frac{\lambda_k}{\hat{\mu}_k+\epsilon}\,\nabla_{\theta}\mathcal{L}_k,
\end{equation}
which can be viewed as \emph{adaptive preconditioning} that rescales each loss term by the inverse of its running magnitude.
Intuitively, if $\mathcal{L}_k$ is currently large in absolute value, its scale estimate $\hat{\mu}_k$ also grows, reducing its effective weight and preventing it from overwhelming other objectives.
Conversely, if $\mathcal{L}_k$ is naturally small (bounded losses or well-optimized terms), normalization avoids vanishing contributions.
As a result, the coefficients $\lambda_k$ become more interpretable: they primarily control \emph{relative importance} rather than compensating for incompatible numeric scales.
This strategy is lightweight (no additional networks or second-order computation) and is especially suitable for UHD-BIQA where mini-batch diversity and distortion heterogeneity can cause large fluctuations across different loss terms.

\subsection{Computational complexity and memory footprint}
\label{sec:complexity}
\label{sec:runtime}
Let $N=G_wG_h$ denote the number of patches (graph nodes), $k$ the neighborhood size in the KNN graph, and $d$ the node embedding dimension.
The computational cost of the graph reasoning stage is dominated by sparse neighborhood aggregation and per-node linear projections.

Assuming a sparse adjacency with $|\mathcal{E}|\approx Nk$ edges, one GCN layer roughly costs
\begin{equation}
\mathcal{O}(|\mathcal{E}|d) + \mathcal{O}(Nd^2)\approx \mathcal{O}(Nk d + Nd^2),
\end{equation}
where the first term corresponds to message passing and the second term corresponds to the feature transformation by $\mathbf{W}\in\mathbb{R}^{d\times d}$.
The memory footprint is mainly from storing node features $\mathcal{O}(Nd)$ and sparse edges $\mathcal{O}(Nk)$.

In UHD-BIQA, the overall runtime is additionally dominated by patch feature extraction, which requires $N$ forward passes through the CNN backbone (batched in practice).
Therefore, $N$ is a key trade-off parameter, and a larger $N$ provides denser evidence while increases both CNN and graph costs.

\subsection{Parameter settings and implementation details}
\label{sec:impl}
Table~\ref{tab:impl_details} lists the default configuration of the proposed framework. 
Unless otherwise specified, all ablations follow the official UHD-IQA train/validation split for hyper-parameter selection and report performance on the held-out test split.
\begin{table}[ht]
\centering
\caption{Default implementation details of \method{}.}
\label{tab:impl_details}
\small
\setlength{\tabcolsep}{4pt}
\begin{tabular}{ll}
\toprule
Component & Settings \\
\midrule
Patch sampling & Patch size $P=256$; grid $(G_w,G_h)=(18,12)$ ($N=216$) \\
Backbone encoder & ResNet-50 \cite{he2016deep} pre-trained on ImageNet \cite{deng2009imagenet} \\
Node embedding & Global average pooled 2048-dim feature $\rightarrow$ linear proj. to $d=512$ \\
Graph construction & $k$NN graph with $k=24$; edge weight via radial basis function kernel (Eq.~\ref{eq:rbf_affinity}) \\
Distance metric & Hybrid spatial--feature distance with $\lambda_w=0.7$ (Eq.~\ref{hwimp}) \\
GCN & $L=3$ residual graph conv layers; mixing coefficient $\alpha=0.55$ \\
Readout & Gated attention pooling + MLP regressor; affine calibration \\
Regularization & Dropout $0.15$; batch normalization in each GCN block \\
Optimization & AdamW; initial learning rate $1\times 10^{-4}$; weight decay $6\times 10^{-5}$ \\
Evaluation & EMA of weights; deterministic two-pass test-time augmentation \\
Hardware & NVIDIA A800 GPU (80GB) \\
\bottomrule
\end{tabular}
\end{table}

We implement \method{} in PyTorch and use sparse graph operations (e.g., via torch-geometric) to scale message passing to hundreds of nodes per image.
To avoid data leakage, all hyper-parameters (grid size, neighborhood $k$, distance mixing, and loss weights) are selected on the training/validation split only.
For reproducibility, we fix random seeds and report averages over repeated runs when the official protocol permits.

\section{Experiments and Results}
\label{sec:exp}

\subsection{The dataset}
\label{sec:dataset}
Experiments are conducted on the UHD-IQA benchmark \cite{hosu2024uhdiqa}, which contains 6073 UHD photographs standardized to a width of 3840 pixels while preserving aspect ratio. Each image is annotated by 20 expert ratings aggregated into a mean opinion score (MOS). The official split includes 4269 training images, 904 validation images, and 900 testing images. Hyper-parameters are tuned on the training/validation sets, MOS values are normalized by the training mean and standard deviation during optimization, and checkpoints are selected by validation SRCC prior to testing.

\subsection{Involved BIQA methods}
\label{sec:compared_methods}
Three categories of BIQA methods are compared. 
The first group includes four DL-based methods, DBCNN \cite{zhang2020dbcnn},  CONTRIQUE \cite{contrique}, HyperIQA \cite{hosu2024uhdiqa} and ARNIQA \cite{arniqa2024}. 
Their performance metric values are cited from the study \cite{hosu2024uhdiqa}. 
The second group contains the results of the six participant teams from the AIM Challenge \cite{hosu2025aim_uhdiqa_lncs}. 
The last group comsists of GRL-based methods, 
such as GraphIQA \cite{sun2022graphiqa}, NLNet-IQA \cite{jia2022nlnet}, 
and AGA-IQA \cite{wang2024adaptive}. 

\subsection{Evaluation metrics}
\label{sec:metrics}
Three standard BIQA metrics, Spearman rank-order correlation coefficient (SRCC), Pearson linear correlation coefficient (PLCC), and root mean square error (RMSE), are employed. Higher SRCC and PLCC values respectively indicate better monotonic and linear agreement with MOS, while lower RMSE values indicate reduced absolute prediction error. 
Metrics are computed as, 
\begin{equation}
A_{ij} = \exp\left(-\frac{(d_{ij}^{hyb})^2}{\tau}\right),  
\label{eq:rbf_affinity}
\end{equation} 
\begin{equation}
\mathrm{PLCC}=\frac{\sum_{i}(y_i-\bar{y})(\hat{y_i}-\bar{\hat{y}})}{\sqrt{\sum_{i}(y_i-\bar{y})^2}\sqrt{\sum_{i}(\hat{y_i}-\bar{\hat{y}})^2}},
\end{equation}
\begin{equation}
\mathrm{RMSE}=\sqrt{\frac{1}{n}\sum_{i=1}^{n}(y_i-\hat{y_i})^2}.
\end{equation}

\subsection{Performance on the UHD-IQA database}
\label{sec:uhd_results}
Table \ref{allUHD} compares representative BIQA methods from three major categories, general DL-based methods, challenge-leading UHD-specific methods, and GRL-based methods, 
on the UHD-IQA database, revealing several important findings. 
\begin{table}[htbp]
  \centering
  \caption{Comparison of different categories of BIQA algorithms on the UHD-IQA database}
    \begin{tabular}{lllll}
    \toprule
    Categories   
            & \multicolumn{1}{l}{Algorithm} 
            & \multicolumn{1}{l}{PLCC} 
            & \multicolumn{1}{l}{SRCC} 
            & \multicolumn{1}{l}{RMSE} \\
    \midrule
    \multicolumn{1}{l}{\multirow{4}[2]{*}{DL-based}} 
        & HyperIQA \cite{hosu2024uhdiqa} & 0.103 & 0.553 & 0.118 \\
        & CONTRIQUE \cite{contrique}     & 0.678 & 0.732 & 0.073 \\
        & ARNIQA \cite{arniqa2024}       & 0.694 & 0.739 & 0.074 \\
        & DBCNN \cite{zhang2020dbcnn}    & 0.723 & 0.745 & 0.154 \\
    \midrule
    \multicolumn{1}{l}{\multirow{5}[2]{*}{The Challenge \cite{hosu2024uhdiqa}}} 
            & SJTU (UIQA) \cite{sun2024assessing} & 0.7985 & \textbf{0.8463} & 0.0615 \\
            & GS-PIQA       & 0.7925 & 0.8297 & 0.0607 \\
            & CIPLAB        & \textbf{0.7995} & 0.8354 & 0.0638 \\
            & EQCNet        & 0.7682 & 0.7954 & 0.0621 \\
            & MobileNet-IQA & 0.7559 & 0.7883 & 0.0659 \\
    \midrule
    \multicolumn{1}{l}{\multirow{3}[2]{*}{GRL-based}} 
            & AGA-IQA \cite{wang2024adaptive} & 0.5914 & 0.6426 & 0.0972 \\ 
            & GraphIQA \cite{sun2022graphiqa} & 0.6420 & 0.6632 & 0.0875 \\
            & NLNet-IQA \cite{jia2022nlnet}   & 0.6804 & 0.7122 & 0.0675 \\     
    \midrule
    GRL-based  & \method{} (ours) & {0.7784} & {0.8019} & \textbf{0.0519} \\
    \bottomrule
    \end{tabular}%
  \label{allUHD}%
\end{table}%

Among DL-based approaches, performance is generally limited on UHD-IQA, 
suggesting that methods originally designed for standard-resolution image quality prediction struggle to generalize to UHD perceptual distortions. 
DBCNN \cite{zhang2020dbcnn} achieves the strongest PLCC (0.723) in this category, 
while CONTRIQUE \cite{contrique} and ARNIQA \cite{arniqa2024} provide relatively competitive SRCC values (0.732 and 0.739, respectively). 
HyperIQA \cite{hosu2024uhdiqa} performs notably poorly, 
indicating substantial instability under UHD conditions. 
Overall, these results suggest that patch-based or DL-based strategies may inadequately capture the complex multi-scale and spatially heterogeneous distortions present in UHD content.

The methods from the UHD-IQA Challenge consistently outperform DL-based baselines, confirming the value of architectures specifically optimized for UHD settings. 
CIPLAB achieves the highest PLCC (0.7995), 
while SJTU (UIQA) \cite{sun2024assessing} obtains the best SRCC (0.8463), 
indicating particularly strong ranking consistency with subjective human perception. 
These methods likely benefit from specialized patch aggregation, 
distortion-aware modeling, or multi-cues driven optimization. 

GRL-based methods \cite{wang2024adaptive, sun2022graphiqa, jia2022nlnet} 
show mixed results. 
Graph modeling is theoretically suitable for preserving 
structural relationships among image regions, 
while existing methods, such as AGA-IQA \cite{wang2024adaptive}, 
GraphIQA \cite{sun2022graphiqa} and NLNet-IQA \cite{jia2022nlnet}, obtain relatively inferior performance to those methods in the Challenge \cite{hosu2024uhdiqa}. 
This suggests that earlier graph-based BIQA approaches 
may not fully exploit UHD-specific spatial dependencies 
or may suffer from suboptimal graph construction and reasoning mechanisms.

Importantly, the proposed method \method{} achieves a strong balance 
across all evaluation metrics, with PLCC = 0.7784, SRCC = 0.8019, 
and the best RMSE = 0.0519 among all compared methods. 
While its PLCC and SRCC values are slightly below the top challenge performers, 
the substantially lower RMSE indicates superior absolute prediction accuracy and more stable quality score estimation. This is a particularly valuable property in practical UHD quality assessment applications, where precise MOS regression is often more critical than rank consistency alone.

From a methodological perspective, these results suggest that the proposed graph-based framework successfully enhances prediction precision by better modeling structured patch relationships and reducing regression error, even if further optimization may still improve ranking monotonic and linear agreement. The performance demonstrates that graph reasoning can be highly competitive for UHD-IQA when carefully designed, and may offer stronger robustness in practical deployment.

\subsection{Ablation studies}
UHD-BIQA imposes practical constraints on memory and computational cost as image resolution increases. 
Under the patch-to-graph formulation, the number of graph nodes is $N = G_w G_h$, 
and the edge number in a KNN graph scales approximately as $\mathcal{O}(Nk)$. 
For message-passing layers, the per-layer computational complexity is roughly $\mathcal{O}(Nk d)$. 
Consequently, increasing either $N$ or $k$ leads to substantially higher computational and memory demands, making both key bottlenecks in UHD-BIQA. 

To keep the ablation design interpretable and computationally tractable, the following studies are organized in a staged manner. 
First, we determine the structural hyper-parameters of graph construction 
under a single objective. 
In other word, we set $\lambda_{\mathrm{mse}}=1$, and  $\lambda_{\mathrm{corr}}=\lambda_{\mathrm{rank}}=\lambda_{\mathrm{var}}=0$ 
in the Table~\ref{tab:gridtrade} before finding the optimal coefficients 
of the full EMA-normalized multi-loss objective. 
This design keeps the graph-topology search consistent with the grid-selection stage and avoids introducing additional uncertainty from the later multi-loss coupling. 
In this way, the selected $k$ reflects a stable graph-building choice rather than a parameter tuned jointly with the full objective.

After the grid size and graph-construction protocol are fixed, 
we investigate the coefficient setting of the full loss 
and the final graph-distance formulation. 
In this way, the effects of graph scale, neighborhood connectivity, 
and optimization design can be examined with reduced mutual interference.

\subsubsection{The sampling grid number under a single training objective}
\label{sec:gridtrade}
Table~\ref{tab:gridtrade} shows the prediction accuracy and training cost under different grid resolutions. 
When the grid number increases from $N=150$ to $N=216$, both SRCC and PLCC improve noticeably, while RMSE decreases from 0.0632 to 0.0602. 
This trend suggests that denser spatial sampling provides richer local evidence and helps the graph model capture more informative distortion patterns in UHD images. 
However, increasing grid density also leads to higher computational cost, as reflected by the longer wall-clock training time.
\begin{table}[H]
\centering
\caption{BIQA performance and time cost regarding the grid size under a simple setting ($\lambda_{\mathrm{mse}}=1$).}
\label{tab:gridtrade}
\small
\setlength{\tabcolsep}{3.5pt}
\begin{tabular}{c|c|c|ccc|c}
\hline
Grid & $N$ & best $k$ & SRCC $\uparrow$ & PLCC $\uparrow$ & RMSE $\downarrow$ & Time (min) $\downarrow$\\
\hline
15$\times$10 & 150 & 20 & 0.7791 & 0.7281 & 0.0632 & 485.8 \\
18$\times$12 & 216 & 24 & \textbf{0.7920} & \textbf{0.7545} & \textbf{0.0602} & 603.0 \\
21$\times$14 & 294 & -- & -- & -- & -- & -- \\
\hline
\end{tabular}
\end{table}

In this controlled setting, the $18\times12$ grid ($N=216$) offers a favorable 
trade-off between prediction performance and computational feasibility, 
providing a relatively strong balance without claiming it as the absolute best. 
When the grid resolution is further increased to $21\times14$ ($N=294$), training becomes infeasible due to out-of-memory errors, indicating that graph-related memory and computation grow too quickly for UHD-BIQA at overly dense sampling resolutions. 
Based on these observations, we adopt the $18\times12$ grid as a balanced and stable configuration for the subsequent experiments, including the search of the graph neighborhood size and the later coefficient optimization of the full multi-loss objective.

\subsubsection{The neighborhood size $k$ for KNN graph construction under a single training objective}
\label{sec:kscan_main}
\begin{figure}[htbp]
  \centering
  \SafeIncludeGraphicsTwo[width=0.78\linewidth]{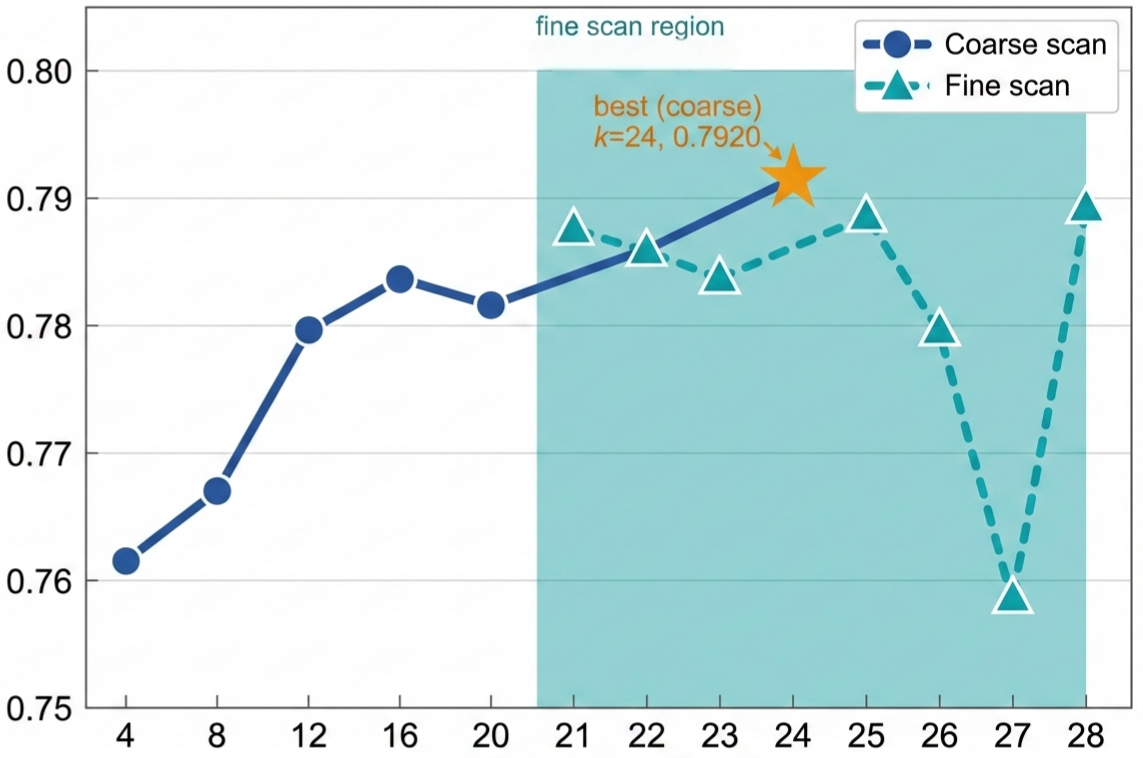}{figs/fine and coarse2.png}
  \caption{Coarse-to-fine search of the neighborhood size $k$ for graph construction.}
  \label{fig:kscan}
\end{figure}

Figure~\ref{fig:kscan} presents a coarse-to-fine search procedure for the neighborhood size $k$. The coarse scan first evaluates a compact candidate set between 4 and 28 with an equal interval of 4, and $k=24$ is identified as the most promising option. 
A subsequent fine-grained examination around this region between 21 and 28 with an equal interval of 1 further confirms that $k=24$ remains the best-performing neighborhood size for graph construction.

The result indicates that a moderately sized local neighborhood is sufficient to capture the structural dependency among sampled regions, since using too few neighbors may lead to incomplete local connectivity, whereas overly large neighborhoods may introduce redundant or less informative relations. 

\subsubsection{Coefficient search for the EMA-normalized multi-loss objective}
\label{sec:loss_ablation}
Under the settings of the grid $ = 18\times12$ and $k = 24 $, 
we investigate the coefficients under the EMA-normalized multi-loss objective 
(Eq.~\eqref{eq:ema_norm_loss}). 
Table~\ref{tab:loss_search} summarizes all tested coefficient settings, including the single-term cases and multi-loss combinations, and sorts them by SRCC in descending order.
\begin{table}[!h]
\centering
\caption{Coefficient search under the EMA-normalized multi-loss objective (sorted by SRCC).}
\label{tab:loss_search}
\small
\setlength{\tabcolsep}{5pt}
\renewcommand{\arraystretch}{1.08}
\begin{tabular}{ccccccc}
\toprule
$\lambda_{\mathrm{corr}}$ & $\lambda_{\mathrm{mse}}$ & $\lambda_{\mathrm{rank}}$ & $\lambda_{\mathrm{var}}$ & PLCC$\uparrow$ & SRCC$\uparrow$ & RMSE$\downarrow$ \\
\midrule
0.8 & 0.0 & 0.2 & 0.0 & 0.7744 & \textbf{0.7982} & 0.0567 \\
0.6 & 0.4 & 0.0 & 0.0 & 0.7548 & 0.7939 & 0.0540 \\
0.0 & 1.0 & 0.0 & 0.0 & 0.7545 & 0.7920 & 0.0602 \\
0.4 & 0.6 & 0.0 & 0.0 & 0.7540 & 0.7919 & 0.0611 \\
0.2 & 0.0 & 0.8 & 0.0 & 0.7626 & 0.7914 & 0.0576 \\
0.2 & 0.8 & 0.0 & 0.0 & 0.7495 & 0.7906 & 0.0601 \\

1.0 & 0.0 & 0.0 & 0.0 & 0.7673 & 0.7897 & 0.0612 \\
0.0 & 0.8 & 0.2 & 0.0 & 0.7439 & 0.7860 & 0.0620 \\
0.8 & 0.2 & 0.0 & 0.0 & 0.7464 & 0.7847 & 0.0536 \\
0.8 & 0.0 & 0.0 & 0.2 & 0.7516 & 0.7828 & 0.0619 \\
0.0 & 0.6 & 0.4 & 0.0 & 0.7476 & 0.7820 & 0.0540 \\
0.0 & 0.4 & 0.6 & 0.0 & 0.7489 & 0.7799 & 0.0574 \\
0.0 & 0.0 & 0.6 & 0.4 & 0.7621 & 0.7780 & 0.0567 \\
0.6 & 0.0 & 0.4 & 0.0 & 0.7465 & 0.7730 & 0.0570 \\
0.0 & 0.0 & 0.8 & 0.2 & 0.7381 & 0.7694 & 0.0584 \\
0.4 & 0.0 & 0.6 & 0.0 & 0.7427 & 0.7678 & 0.0568 \\
0.0 & 0.8 & 0.0 & 0.2 & 0.7160 & 0.7508 & 0.0455 \\
0.0 & 0.0 & 1.0 & 0.0 & 0.7220 & 0.7426 & 0.0594 \\
0.0 & 0.2 & 0.8 & 0.0 & 0.7565 & 0.7201 & 0.0599 \\
0.0 & 0.0 & 0.4 & 0.6 & 0.6410 & 0.6591 & 0.0601 \\
0.6 & 0.0 & 0.0 & 0.4 & 0.6158 & 0.6476 & 0.0575 \\
0.4 & 0.0 & 0.0 & 0.6 & 0.3670 & 0.3955 & 0.0572 \\
0.0 & 0.0 & 0.2 & 0.8 & 0.2904 & 0.3363 & 0.0573 \\
0.0 & 0.4 & 0.0 & 0.6 & 0.1384 & 0.2932 & 0.0639 \\
0.2 & 0.0 & 0.0 & 0.8 & 0.2283 & 0.2847 & 0.0578 \\
0.0 & 0.2 & 0.0 & 0.8 & 0.1475 & 0.2614 & 0.0630 \\
0.0 & 0.0 & 0.0 & 1.0 & 0.1316 & 0.2421 & 0.0662 \\
0.0 & 0.6 & 0.0 & 0.4 & 0.0635 & 0.2227 & 0.0640 \\
\bottomrule
\end{tabular}
\end{table}

It is found that $\lambda_{\mathrm{corr}}=0.8$, $\lambda_{\mathrm{mse}}=0.0$, $\lambda_{\mathrm{rank}}=0.2$, and $\lambda_{\mathrm{var}}=0.0$ 
lead to the highest SRCC value (0.7982). 
Several mixed settings also outperform the single-term cases, indicating that correlation and ranking objectives can complement score regression when the coefficients are chosen appropriately. 
By contrast, configurations dominated by the variance term ($\lambda_{\mathrm{var}}=1.0$ ) lead to clear performance degradation. 

The observations suggest that the EMA-normalized formulation is beneficial, 
while its effectiveness depends on careful coefficient selection. 
The best-performing coefficient setting is therefore adopted in the subsequent graph-topology study, where the graph scale and neighborhood size remain fixed so that the effect of the graph distance measure can be examined in a controlled manner.

\subsubsection{Effect of different distance measures in graph construction}
\label{sec:distance_measure}
The graph topology is evaluated under various distance metrics 
and weighting schemes (Eq.~\ref{hwimp}). 
As shown in Table~\ref{tab:distance_measure}, we compare feature-based, 
Euclidean-based, and hybrid KNN graph constructions. 
The results consistently demonstrate that spatial proximity is 
the primary factor governing the effectiveness of graph representation. 
\begin{table}[!h]
\centering
\caption{Effect of graph distance measures under different weights $\lambda_w$.}
\label{tab:distance_measure}
\small
\setlength{\tabcolsep}{6pt}
\renewcommand{\arraystretch}{1.12}
\begin{tabular}{lcccc}
\toprule
Measure & $\lambda_w$ & PLCC$\uparrow$ & SRCC$\uparrow$ & RMSE$\downarrow$ \\
\midrule
Feature-only (semantic)   & 0.0 & 0.7216          & 0.7434          & 0.0728          \\
Hybrid                    & 0.1 & 0.7466          & 0.7683          & 0.0672          \\
Hybrid                    & 0.3 & 0.7682          & 0.7894          & 0.0593          \\
Hybrid                    & 0.5 & 0.7741          & 0.7933          & 0.0598          \\
Hybrid                    & 0.7 & \textbf{0.7784} & \textbf{0.8019} & \textbf{0.0519} \\
Hybrid                    & 0.9 & 0.7709          & 0.7915          & 0.0617          \\
Euclidean-only (spatial)  & 1.0 & 0.7744          & 0.7982          & 0.0567          \\
\bottomrule
\end{tabular}
\end{table}

As shown in Table \ref{tab:distance_measure}, incorporating spatial distance into graph construction significantly improves performance over the feature-only baseline. 
The hybrid graph exhibits a consistent performance gain as $\lambda_w$ increases 
from 0.1 to 0.7, indicating that spatial information effectively 
regularizes graph connectivity and enhances perceptual consistency. 
The best performance is achieved at $\lambda_w=0.7$, suggesting 
an optimal balance between spatial proximity and feature similarity. 
Further increasing $\lambda_w$ leads to performance degradation, 
implying that excessive reliance on spatial structure suppresses discriminative feature relationships. 
Notably, the spatial-only graph ($\lambda_w=1.0$) remains competitive 
but does not surpass the optimal hybrid setting, 
demonstrating that spatial information is more important for graph construction.

\subsection{Computing complexity}
\label{sec:computing_complexity_results}
Table~\ref{tab:computing_complexity} compares the computational complexity of current UHD-BIQA frameworks in terms of input size, parameter count, and multiply--accumulate operations (MACs). 
The proposed framework has the smallest model size, indicating strong parameter efficiency, but it also requires substantially higher MACs. 
\begin{table}[htbp]
\centering
\caption{Comparison of computing complexity across current UHD-BIQA frameworks.}
\label{tab:computing_complexity}
\small
\setlength{\tabcolsep}{4pt}
\renewcommand{\arraystretch}{1.12}
\resizebox{\textwidth}{!}{%
\begin{tabular}{lccc}
\toprule
Method & Input size & Params. (M) & MACs (G) \\
\midrule
SJTU (UIQA)~\cite{sun2024assessing} 
    & $480\times480$      & 82.85     & 43.53     \\
GS-PIQA~\cite{hosu2025aim_uhdiqa_lncs} 
    & $384\times384$      & 144.81    & 50.26     \\
\method{} (ours) 
    & $(18\times256)\times(12\times256)$ 
                           & 25.61     & 1153.59   \\
\bottomrule
\end{tabular}%
}
\end{table}

It is found that \method{} offers the most parameter-efficient design 
among the approaches, requiring only 25.61 M parameters 
compared with 82.85M for SJTU and 144.81M for GS-PIQA. 
This lightweight model size reduces storage demands and simplifies deployment. 
However, these advantages come at the cost of substantially higher computational complexity, with \method{} requiring 1153.59 GMACs, far exceeding the 43.53G and 50.26G required by the compared methods. 
This heavy computational burden is mainly caused by its multi-patch encoding and graph reasoning strategy, which enhances detailed perceptual modeling for UHD images but significantly increases inference cost. 
Overall, \method{} achieves a favorable balance in model compactness and training simplicity, but its high MACs indicate that further optimization 
is needed for real-time or resource-constrained applications.

\section{Discussion}
\label{sec:discussion}
UHD-BIQA is challenging due to the extremely high spatial resolution of input images and the limited computational resources available in practical systems. 
To address this challenge, we propose the \method{} framework that designs aspect-ratio-aligned patch sampling, hybrid KNN graph construction, and EMA-normalized optimization of a hybrid loss function (Figure \ref{fig:overview}). 
Extensive experimental results suggest that 
the proposed framework achieves promising performance 
on the UHD-IQA database (Table \ref{allUHD}), 
and the performance is obtained based on the grid patch 
number $N = 18 \times 12$ (Table~\ref{tab:gridtrade}), 
$k = 24$ in KNN graph building (Figure~\ref{fig:kscan}), 
$\lambda_{corr} = 0.8$ and $\lambda_{rank} = 0.2$ in 
the hybrid loss function (Eq.~\ref{loss}), 
and $\lambda_{w} = 0.7$ in the weighted distance metric (Eq.~\ref{hwimp}). 

\subsection{Why graph reasoning is suitable for UHD-BIQA}
A practical advantage of graph reasoning for UHD-BIQA is that it places all sampled patches into a unified relational structure instead of treating them as isolated crops. 
This matters because UHD quality is rarely determined by a single local artifact alone; rather, local degradations are interpreted in the context of surrounding regions and the broader scene layout. 
Resize-and-crop pipelines can preserve some local detail, but they usually fuse different views only after separate feature extraction \cite{hosu2025aim_uhdiqa_lncs, gu2025super, sun2024assessing}. 
By contrast, graph message passing provides an explicit mechanism for contextual interaction among all sampled regions, and gated readout lets the final prediction emphasize the most informative regions without discarding scene-level context \cite{sun2022graphiqa, jia2022nlnet, wang2024adaptive, liu2025iqg}.

The experimental results are consistent with this interpretation. 
Table~\ref{tab:gridtrade} shows that performance improves when the sampling grid becomes dense enough to expose richer local evidence, and Figure~\ref{fig:kscan} shows that graph reasoning is most effective only when the neighborhood size is chosen carefully rather than enlarged arbitrarily. 
Table~\ref{allUHD} further shows that the resulting pipeline achieves the best RMSE among the public UHD-oriented methods. 
Taken together, these observations support the basic motivation behind graph reasoning for UHD-BIQA that local evidence is useful, and it becomes more reliable when it is embedded in a structured global context rather than pooled independently. 
This is also consistent with scale-aware UHD studies \cite{huang2024hriq, hosu2025iisa}, which suggest that perceptual quality is entangled with native spatial structure.

\subsection{Why \method{} does not outperform the Challenge champion}
Although the graph formulation is effective, 
the proposed framework does not outperform the top-tier methods, 
such as SJTU/UIQA \cite{sun2024assessing} 
and GS-PIQA \cite{hosu2025aim_uhdiqa_lncs}. 
The reasons are manifold. 
Generally, SJTU/UIQA combines global aesthetics, local distortion fragments, 
and saliency-guided regions through three coordinated 
Transformer branches \cite{sun2024assessing}. 
GS-PIQA further enhances ranking-oriented perceptual modeling 
by incorporating grid mini-patch sampling, pyramid perception, 
and a larger Swin-base backbone under the challenge setting \cite{hosu2025aim_uhdiqa_lncs}. 
These models achieve superior performance by leveraging advanced backbone architectures, enriched feature representations, and effective multi-scale semantic integration.

The proposed \method{} framework achieves slightly inferior correlation performance while attaining the lowest prediction error and the smallest parameter count, although it requires substantially higher MACs than the compared UHD-oriented methods (Tables~\ref{allUHD} and~\ref{tab:computing_complexity}).

The effectiveness of the framework can be attributed to several factors.
First, the framework unifies patch encoding, graph construction, and interactive message passing within a single structured pipeline.
This design facilitates more accurate absolute score calibration, as evidenced by the best RMSE reported in Table \ref{allUHD}.
Second, \method{} adopts a pre-trained ResNet-50 \cite{he2016deep} as the feature extractor, which is computationally lightweight but less powerful than the more advanced backbone architectures used in SJTU/UIQA and GS-PIQA \cite{hosu2025aim_uhdiqa_lncs}.
Integrating hundreds of patches enhances detailed perceptual representation for UHD content, while it also causes extremely high MACs (Table~\ref{tab:computing_complexity}). 
Overall, \method{} represents a balanced trade-off between explicit relational reasoning and stronger multi-branch feature extraction for addressing the challenges of UHD-BIQA at the cost of heavier training cost and reduced real-time suitability.

\subsection{Limitations of the current study}
Several limitations should be acknowledged in the current study.
A major practical constraint of the proposed framework lies in its computational cost. 
As reported in Table~\ref{tab:computing_complexity}, 
\method{} contains a relatively small number of learnable parameters, 
while its patch-wise encoding of UHD images and the subsequent graph message passing still lead to high MACs. 
The grid-size analysis in Table~\ref{tab:gridtrade} further indicates that denser patch sampling can improve prediction accuracy, but this benefit comes with a rapid increase in training cost. 
These results suggest that the current framework may still be less suitable for real-time or resource-constrained UHD-IQA applications. 
Beyond efficiency, the representational capacity of the current model also remains limited when compared with stronger UHD-specific Transformer-based or multi-branch systems \cite{hosu2025aim_uhdiqa_lncs}. 
As shown in Table~\ref{allUHD}, \method{} achieves the lowest RMSE among the compared methods, but its PLCC and SRCC are still lower than those of the top challenge methods \cite{sun2024assessing, hosu2025aim_uhdiqa_lncs}. 
This observation suggests that the ResNet-50 backbone \cite{he2016deep} and the GCN module may not fully capture richer multi-scale semantic, aesthetic, and saliency-aware cues that are exploited by heavier architectures. 
Meanwhile, graph construction requires further investigation. 
The ablation studies show that the final performance is affected by several coupled factors, including the sampling grid, the neighborhood size $k$, the multi-loss coefficients, and the spatial-feature mixing weight $\lambda_w$. 
Owing to the high computational cost, these factors are examined in a staged manner rather than through exhaustive joint optimization. 
Finally, the scope of the experimental validation is limited. 
The main evaluation is conducted on the UHD-IQA benchmark \cite{hosu2024uhdiqa}, and broader validation on additional databases is needed to better examine cross-dataset generalization.

\section{Conclusions}
\label{sec:conclusion}
This paper presents a graph-based framework, \method{}, for UHD-BIQA, aiming to alleviate the challenges associated with extremely high resolution and global--local inconsistency.
The framework combines aspect-ratio-aligned image patching, KNN-based graph construction, and multi-objective loss weighting to predict image-level quality scores by propagating information among local patch representations through GCN-based global--local message passing.
On the UHD-IQA benchmark, \method{} obtains the lowest RMSE, suggesting its effectiveness in reducing prediction error.
However, its overall performance remains less competitive than recent multi-branch architectures and Transformer-based methods, indicating that there is still room for improvement in feature representation and graph modeling.
Future work will explore more efficient patch encoding, adaptive graph construction, stronger visual backbones, and more comprehensive evaluation on additional UHD image quality datasets.

\newpage
\appendix
\section{Reference implementation sketch}
\label{app:pseudo}
Table~\ref{alg:inference} summarizes the core inference pipeline of \method{} at a high level. It can serve as a concise reproducibility checklist when re-implementing the method.

\begin{table}[!h]
\centering
\caption{High-level inference pipeline of \method{}.}
\label{alg:inference}
\begin{tabular}{p{0.95\linewidth}}
\toprule
\textbf{Input:} UHD image $\mathbf{I}$; grid $(G_w,G_h)$; patch size $P$; neighborhood size $k$ \\
\textbf{Output:} predicted quality $\hat{y}$ \\
\midrule
1. Sample $N = G_w G_h$ patch centers on an aspect-ratio--aligned grid. \\
2. Extract patches $\{\mathbf{x}_i\}_{i=1}^{N}$ of size $P\times P$ from $\mathbf{I}$. \\
3. Encode each patch with a CNN backbone to obtain embeddings $\{\mathbf{h}_i\}$. \\
4. Build the graph using the general hybrid KNN formulation with the selected spatial-feature mixing weight. \\
5. Apply $L$ residual GCN layers to propagate context on the graph. \\
6. Compute attention weights $\{a_i\}$ and aggregate $\mathbf{z}=\sum_i a_i\mathbf{h}_i$. \\
7. Predict $\hat{y}=s f(\mathbf{z})+b$ with affine calibration. \\
\bottomrule
\end{tabular}
\end{table}

\vspace{6pt}

\authorcontributions{Conceptualization, S.Y. and Q.S.; methodology, S.Y.; software, E.C., M.H. and X.R.; validation, E.C., M.H. and X.R.; formal analysis, S.Y., E.C. and S.Z.; investigation, S.Y. and E.C.; resources, Q.S. and Z.Z.; data curation, S.Y.; writing---original draft preparation, S.Y.; writing---review and editing, S.Y., E.C., Z.Z. and Q.S.; visualization, M.H., X.R. and S.Z.; supervision, Q.S. and Z.Z.; project administration, Q.S. and Z.Z. All authors have read and agreed to the published version of the manuscript.}

\funding{The work was in part supported by the National Key Research and Develop Program of China (Grant No. 2024YFF0907401, 2022ZD0115901, and 2022YFC2409000), the National Natural Science Foundation of China (Grant No. 62177007, U20A20373, and 82202954), the China-Central Eastern European Countries High Education Joint Education Project (Grant No. 202012), the Application of Trusted Education Digital Identity in the Construction of Smart Campus in Vocational Colleges (Grant No. 2242000393), the Knowledge Blockchain Research Fund (Grant No. 500230), and the Medium- and Long-term Technology Plan for Radio, Television and Online Audiovisual (Grant No. ZG23011). The work was also supported by Public Computing Cloud, CUC. The funders had no role in the design of the study; in the collection, analyses, or interpretation of data; in the writing of the manuscript; or in the decision to publish the results.}

\institutionalreview{Not applicable.}

\informedconsent{Not applicable.}

\dataavailability{The UHD-IQA benchmark database used in this study is available via official release.}

\conflictsofinterest{The authors declare no conflict of interest.}



\clearpage
\abbreviations{Abbreviations}{%
The following abbreviations are used in this manuscript.\\

\noindent
\begingroup
\small
\renewcommand{\arraystretch}{0.92}
\begin{tabular}{@{}ll}
AIM & Advances in Image Manipulation \\
BIQA & Blind Image Quality Assessment \\
CNN & Convolutional Neural Network \\
DL & Deep Learning \\
EMA & Exponential Moving Average \\
GCN & Graph Convolutional Network \\
GNN & Graph Neural Network \\
GPU & Graphics Processing Unit \\
GRL & Graph Representation Learning \\
IQA & Image Quality Assessment \\
$k$NN & $k$-Nearest Neighbor \\
MAC & Multiply--Accumulate \\
MOS & Mean Opinion Score \\
MSE & Mean Squared Error \\
PLCC & Pearson Linear Correlation Coefficient \\
ReLU & Rectified Linear Unit \\
RMSE & Root Mean Square Error \\
SRCC & Spearman Rank-Order Correlation Coefficient \\
UHD & Ultra-High Definition \\
UHD-BIQA & Ultra-High-Definition Blind Image Quality Assessment \\
UHD-IQA & Ultra-High-Definition Image Quality Assessment \\
\end{tabular}
\endgroup
}

\begin{adjustwidth}{-\extralength}{0cm}
\reftitle{References}

\begingroup
\small
\sloppy
\setlength{\emergencystretch}{3em}
\urlstyle{same}

\endgroup

\end{adjustwidth}

\end{document}